\pdfoutput=1
\documentclass[conference,compsoc]{IEEEtran}
\IEEEoverridecommandlockouts

\usepackage{tikz}
\usetikzlibrary{positioning, arrows.meta}
\usepackage{listings}
\usepackage{graphicx}
\usepackage{booktabs}
\usepackage{xcolor}

\ifCLASSOPTIONcompsoc
  \usepackage[nocompress]{cite}
\else
  \usepackage{cite}

\fi

\usepackage{booktabs}
 \usepackage{amsmath}
\usepackage{amssymb}
\usepackage{multirow}

\usepackage{url}

\usetikzlibrary{positioning}
\usetikzlibrary{arrows.meta}
\usetikzlibrary{shapes}
\usetikzlibrary{backgrounds}
\makeatletter
\def\@IEEEfigurecaptionsepspace{\vskip 4pt plus 1pt minus 1pt\relax}
\def\@IEEEtablecaptionsepspace{\vskip -2pt\relax}
\renewcommand\section{\@startsection{section}{1}{\z@}{0.7\baselineskip plus 0.2\baselineskip minus 0.1\baselineskip}{0.4\baselineskip plus 0.1\baselineskip}{\normalfont\large\bfseries}}
\renewcommand\subsection{\@startsection{subsection}{2}{\z@}{0.6\baselineskip plus 0.2\baselineskip minus 0.1\baselineskip}{0.3\baselineskip plus 0.1\baselineskip}{\normalfont\sublargesize\bfseries}}
\makeatother

\usepackage{eso-pic}
\AddToShipoutPictureFG*{%
  \AtPageLowerLeft{%
    \put(\LenToUnit{0.5\paperwidth},\LenToUnit{0.30in}){%
      \makebox[0pt][c]{\parbox[b]{\textwidth}{\centering\footnotesize
        Accepted at the 6th Workshop on Knowledge Graphs and Big Data (KGBigData 2026), held with IEEE BigData 2026.\\
        \copyright~2026 IEEE. Personal use of this material is permitted. Permission from IEEE must be obtained for all other uses, in any current or future media, including reprinting/republishing this material for advertising or promotional purposes, creating new collective works, for resale or redistribution to servers or lists, or reuse of any copyrighted component of this work in other works.}}}}}

\begin{document}
\title{%
Neuro-Symbolic Reasoning for Electronics Supply Chain Knowledge Graphs}

\author{\IEEEauthorblockN{Fatemeh Elhambakhsh}
\IEEEauthorblockA{%
Arizona State University\\
Tempe, Arizona\\
Email: felhamba@asu.edu}
\and
\IEEEauthorblockN{Arquimedes Canedo}
\IEEEauthorblockA{\textit{SIEMENS}\\
Princeton, New Jersey\\
Email: arquimedes.canedo@siemens.com}
\and
\IEEEauthorblockN{Arun Ramamurthy}
\IEEEauthorblockA{\textit{SIEMENS}\\
Princeton, New Jersey\\
Email: arun.ramamurthy@siemens.com}
\and
\IEEEauthorblockN{Hyunjee Jin}
\IEEEauthorblockA{\textit{SIEMENS}\\
Princeton, New Jersey\\
Email: hyunjee.jin@siemens.com}
\and
\IEEEauthorblockN{Aleksandar Bradic}
\IEEEauthorblockA{\textit{SIEMENS}\\
New York City, New York\\
Email: aleksandar.bradic@siemens.com}
\and
\IEEEauthorblockN{Hyunwoong Ko\IEEEauthorrefmark{1}}
\IEEEauthorblockA{%
Arizona State University\\
Tempe, Arizona\\
Email: hyunwoong.ko@asu.edu }
\thanks{\IEEEauthorrefmark{1}Corresponding author: Hyunwoong Ko (hyunwoong.ko@asu.edu).}
}

\maketitle

\begin{abstract}
Electronics procurement asks whether a part will be in stock, whether an
offer will keep an active lifecycle status and whether searches for a part will exceed a
threshold in the coming weeks. The relevant data are
large tables of part searches and offer listings, observed sparsely, in which
most states persist between observations. We construct a temporal property
graph of component availability from 632 million search records, with
billions of versioned offer states and daily search edges, and read it as
thresholded temporal atoms: a stock or lifecycle state of a part or offer on a
target day, or a search count over a window, as seen from a weekly anchor. We forecast the unobserved atoms with a temporal extension of a neural Markov
logic network whose rules, recurrent graph encoder and change-weighted
objective are defined over time. On a slice with 0.4 million entities and 178
million temporal facts, one inference pass scores the 32.7 million atoms of
three held-out weeks in 16.5 seconds. We evaluate three tasks separately. In classification of all test atoms, the last observation agrees with more
than 90 percent of held-out labels, so a persistence reference stays slightly
ahead on discrimination, while our model has less than half its negative log-likelihood. In filtered ranking, with ties counted against the scorer, our model doubles the mean reciprocal rank of persistence. In the detection of changes that the last observation misses, such as a stockout, measured on a sample enriched for errors
of that reference, our model reaches an AUC-ROC of 0.61 on these errors against 0.55 for a large language model agent querying the graph with Cypher and 0.51 for a raw value rule, while the agent is more accurate
where the reference is right.
\end{abstract}

\IEEEpeerreviewmaketitle

\section{Introduction}

Electronic equipment manufacturers source thousands of components through a
network of distributors, and the unavailability of a single component can halt
production. Procurement therefore asks recurring questions: which distributors
hold a component and in what quantity, whether a listing will still be there
next month, and how demand for the part is developing. Distributors publish
stock in their own catalogs, and part-search aggregators collect these
catalogs on behalf of design and procurement engineers. Observed over time,
the aggregated responses form a large evolving graph of parts, manufacturers,
distributors and search queries whose edges carry time-varying stock levels,
lifecycle status and search demand. Although central to component selection and
shortage response~\cite{liu2026}, this information has rarely been studied as a
dataset in its own right; prior supply chain graphs rely on firm-level
relations reconstructed from filings, news or language
models~\cite{kosasih2022machine, mungo2024, kose2026}, and the one temporal
supply chain benchmark covers a single firm~\cite{wasi2024supplygraph}.

Such data differ from the curated graphs on which temporal knowledge graph
methods are developed. They are \emph{sparse}: demand concentrates on a few
commodity parts, while most parts have one distributor and few searches. They are \emph{partially observable}: stock is
recorded only when a part is searched for, so the time of a change is never
known exactly. They are \emph{persistent}: most states hold for weeks, so
the informative facts are the few whose value changes. Availability is thus a
slowly changing latent state measured irregularly, not a stream of fully
observed events as in ICEWS~\cite{boschee2015}, GDELT~\cite{leetaru2013} or TGB~\cite{huang2023}.

In current practice these questions are answered by querying the graph, directly or through a language model that writes the query. A query
returns the last recorded state, and a rule can extend it by one step, but neither estimates the states of many offers not observed recently, combines
several weak indications into a graded belief, or forecasts beyond the last fact in the graph. Learned models can address these limits, but
obsolescence and demand forecasters treat each component in
isolation~\cite{jennings2016forecasting, saad2025, moon2022b}, and temporal
knowledge graph models assume dense event streams and represent facts as point
interactions; applying them to component availability requires handling persistent states, irregular observations and quantity thresholds.

We therefore cast the procurement questions as probabilistic reasoning over a
temporal knowledge graph in which each question is a temporal atom: a
predicate holds for an entity on a target day as seen from an anchor day. We
build on the neural Markov logic network (MLN) of
ExpressGNN~\cite{zhang2020expressgnn}, which combines first-order rules with
a graph neural encoder of the variational posterior, and make time explicit
in its grounding, encoder, posterior and objective.
The contributions are threefold. First, we construct a large
temporal knowledge graph of electronic component availability, in which
billions of versioned offer states and daily search edges describe stock,
lifecycle status and search demand observed sparsely and at irregular
intervals. Second, we propose a temporal extension of a neural
MLN~\cite{zhang2020expressgnn} that scales to this graph, with temporal
grounding over the observed support, a recurrent per-anchor encoder, time and
horizon encodings, a change-weighted likelihood and a temporal mirror term.
Third, we evaluate the model over three held-out weeks against persistence and
prior predictors, a large language model (LLM) agent that answers the same
questions by querying the graph with Cypher, and a rule-based scorer, and show
through ablations which component addresses which property of the data.

\section{Related Work}
\label{sec:related}

Knowledge graphs have integrated heterogeneous supply chain sources into one
relational structure, from supply chain maps extracted from
news~\cite{wichmann2020extracting} and resilience
analysis~\cite{liu2023knowledge} to semiconductor-specific graphs for
planning~\cite{herding2025ontology}, pricing~\cite{ramzy2021knowgraphpm} and
retrieval-augmented question answering~\cite{ham2026design}. These graphs are
static snapshots used through queries or language-model retrieval~\cite{elhambakhsh2025domain} and cannot
estimate facts that are absent or lie beyond the last observation. Predictive
work on supply chain graphs has concentrated on firm-level link prediction,
from topological features~\cite{brintrup2018predicting} to graph neural
networks~\cite{kosasih2022machine, aziz2021data, kosasih2024towards,
wasi2024supplygraph}. Closest to our work, Kosasih and
Brintrup~\cite{kosasih2025trustworthy} combine a GNN with logical rules of
fixed confidence for firm-level link prediction, but their graph is static,
the rules do not range over time, and the target is whether a relation exists
rather than which value a state takes on a future day. Obsolescence
forecasting~\cite{sandborn2011forecasting, jennings2016forecasting} predicts
risk from part attributes but treats each part in isolation. No prior work
models the joint evolution of offers, stock and demand across parts as a
temporal graph.

Temporal KG models attach timestamps to facts and either interpolate missing
ones with time-dependent embeddings~\cite{garciaduran2018learning,
lacroix2020tntcomplex} or forecast the next snapshot from past
ones~\cite{jin2020renet, li2021regcn}, and TLogic~\cite{liu2022tlogic} learns
temporal rules for explainable forecasting. All are developed on event
datasets in which each fact is a discrete interaction observed when it occurs
and the set of facts at each timestamp is complete, whereas availability is a
persistent state sampled irregularly and the targets are threshold conditions
on quantities. MLNs~\cite{richardson2006mln} ground weighted first-order rules over a KG and
infer a probability for every atom, but exact inference does not scale to
millions of entities, and rule mining systems~\cite{meilicke2019anyburl} apply
mined Horn rules with fixed confidences but cannot reach atoms no rule covers.
Neural LP~\cite{yang2017neurallp} learns rules differentiably,
pLogicNet~\cite{qu2019plogicnet} alternates an MLN with an embedding model, and
ExpressGNN~\cite{zhang2020expressgnn} parametrizes the variational posterior of
an MLN with a graph neural network. All are evaluated on static benchmarks and
do not handle rules over time, states that persist between observations, or
hundreds of millions of ground atoms.

Three gaps match our contributions: electronics supply chain KGs are static and queried rather than forecast; temporal KG methods assume
dense, complete event streams, while probabilistic logic has not been applied to temporal, sparsely observed data at this
scale; and no prior evaluation places a learned model beside the querying
practice it is meant to replace, with or without a language model.

\section{Temporal Knowledge Graph of the Electronics Supply Chain}\label{sec:data}

\subsection{Electronics Supply Chain Dataset}\label{sec:source}
We sample a subset of data from an electronic component aggregator,
collected between December 20, 2025 and April 20, 2026.
A record contains the user-entered part number, query time and each distributor's matching part entries. An entry
carries the manufacturer name, the manufacturer part number (MPN), a stock
quantity, an optional lifecycle code and a distributor-reported update time. A
record induces a small bipartite structure of one query, about six distributors
and about 25 entries on average, with heavy tails. Table~\ref{tab:months} gives
the volume: 632 million records with 15.7 billion entries from 109 distributor
identifiers. Three properties of the data shape the graph and the method.

\begin{figure}[t]
\centering\includegraphics[width=\columnwidth]{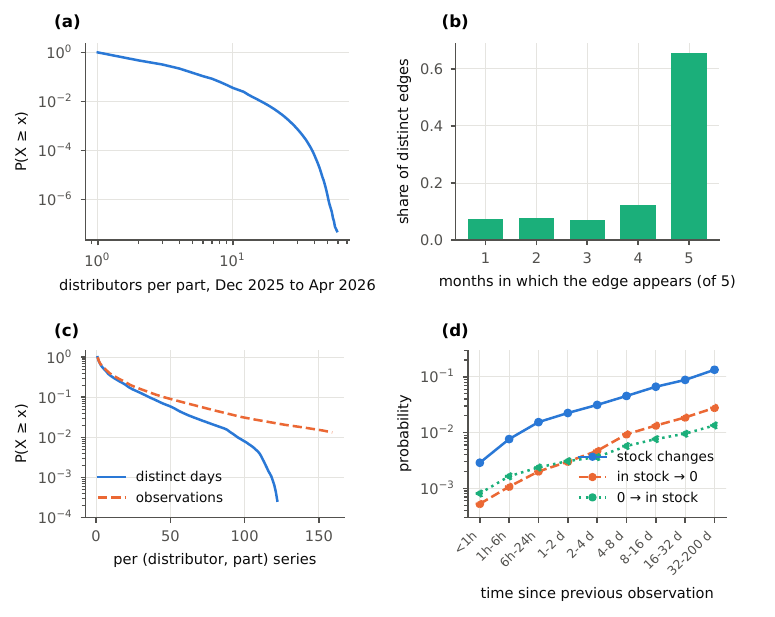}
\caption{The distributor--part data. (a) Distributors per part,
complementary CDF. (b) Share of distributor--part pairs by number of months observed. (c) Distinct days and observations per
series, complementary CDF, from a 1/256 query-sampled panel. (d) Probability
that stock changes, falls to or recovers from zero, by time since the
previous observation.}\label{fig:data}
\end{figure}

\emph{The data are sparse.} The collection is query driven, so a
distributor--part pair is observed only on days when some user searched for the part. The graph contains 39.2 million parts and 104.3 million
distributor--part pairs, a density of 2.4\,\% over 109 distributors; a part has 2.7 distributors on average and more than half of the parts by a single one (Fig.~\ref{fig:data}a), while the ten largest distributors account for about two thirds of all pairs. The support is stable across months
(Fig.~\ref{fig:data}b), so the information lies in the states of the pairs rather than in the pairs themselves.

\emph{The data are partially and irregularly observed.} Most distributor--part
series are observed on a handful of distinct days, few on fifty or more
(Fig.~\ref{fig:data}c). The query time usually lies within an hour of the distributor's update time, so the latter is a usable timestamp for the state.

\emph{States persist.} The probability that the stock value differs from the
previous observation grows with the gap between observations, from below one
percent within a day to above ten percent after a month
(Fig.~\ref{fig:data}d), and transitions into and out of zero stock are an
order of magnitude rarer than changes in quantity. A model must therefore represent a state that persists between observations and assign it a probability on days when it was not observed.

\begin{table}[t]
\caption{Volume of the source data per month. * partial month. Queries and
Parts are distinct normalized queried and returned part numbers within the
month.}\label{tab:months}
\centering\footnotesize
\begin{tabular}{lrrrr}
\toprule
Month & Records (M) & Entries (B) & Queries (M) & Parts (M)\\
\midrule
Dec 2025* & 58.3 & 1.48 & 16.1 & 28.1\\
Jan 2026 & 145.2 & 3.51 & 22.3 & 31.6\\
Feb 2026 & 131.0 & 3.52 & 18.9 & 31.2\\
Mar 2026 & 176.6 & 4.36 & 20.2 & 31.9\\
Apr 2026* & 120.9 & 2.85 & 18.5 & 31.4\\
\midrule
All & 632.0 & 15.71 & & \\
\bottomrule
\end{tabular}
\end{table}

\subsection{Temporal Knowledge Graph Construction}\label{sec:kg}
\label{sec:schema}\label{sec:atoms}\label{sec:construction}
\begin{figure}[t]
\centering
\begin{tikzpicture}[
    node distance=8mm and 8mm,
    ent/.style={draw, rounded corners, fill=blue!8, minimum height=6mm, inner sep=2pt,
                font=\scriptsize, align=center},
    st/.style={ent, fill=orange!15},
    rel/.style={-{Latex[length=2mm]}, font=\tiny, inner sep=1pt}]
  \node[ent] (offer) {Offer\\\{id\}};
  \node[ent, left=9.5mm of offer] (dist) {Distributor\\\{id, name,\\authorized\}};
  \node[ent, right=of offer] (part) {Part\\\{mpn,\\description\}};
  \node[ent, right=10.5mm of part] (man) {Manufacturer\\\{name\}};
  \node[st, below=of offer] (state) {OfferState\\\{stock, lifecycle,\\validFrom, validTo\}};
  \node[st, below=of part, xshift=4mm] (query) {Query\\\{text,\\nSearches, nDays\}};
  \draw[rel] (dist) -- node[above]{OFFERS} (offer);
  \draw[rel] (offer) -- node[above]{FOR} (part);
  \draw[rel] (part) -- node[above]{MADE\_BY} (man);
  \draw[rel] (offer) -- node[left]{HAS\_STATE} (state);
  \draw[rel] ([yshift=2mm]state.west) .. controls +(-8mm,0) and +(-8mm,0) ..
    node[left, align=right]{PREVIOUS\\(newer $\to$ older)} ([yshift=-2mm]state.west);
  \draw[rel] (query.north) -- node[right, align=left]{SEARCHED\_FOR\\\{day, n, exact\}} (query.north |- part.south);
\end{tikzpicture}
\caption{Temporal KG schema. Blue nodes form the static support; dated orange offer states carry supply, and dated \textsc{searched\_for} edges from queries carry demand. \textsc{offers} and \textsc{has\_state} are one-to-many from source to
target, \textsc{previous} is one-to-one, \textsc{searched\_for} is many-to-many,
and the remaining edges are many-to-one.}\label{fig:ontology}
\end{figure}
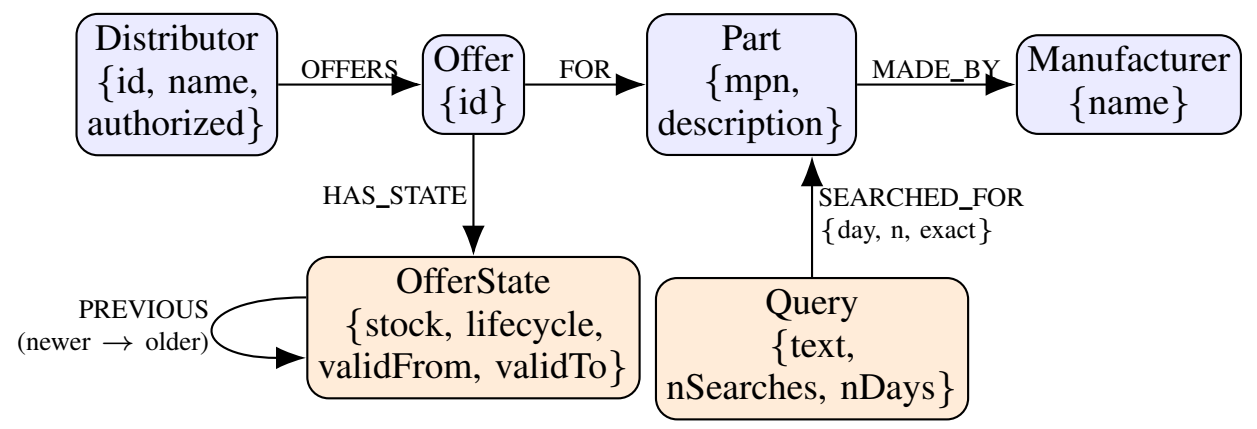

We use a labeled property graph (Fig.~\ref{fig:ontology})
with two design choices. An attribute belongs to the entity whose state it describes: stock and lifecycle are reported per distributor and part and so belong to the distributor--part pair. Time is
represented in the graph itself, so the same graph serves querying and learning. The schema has six node types. \textbf{Manufacturer}(name) and \textbf{Part}(mpn, description) are connected by \textsc{made\_by}; a part is identified by the normalized MPN (upper-cased, non-alphanumeric characters removed), and parts without a manufacturer record have no \textsc{made\_by} edge. A \textbf{Distributor}(id, name, authorized) is linked by \textsc{offers} to an \textbf{Offer}(id), one node per distributor--part pair, which is linked by \textsc{for} to the part. An \textbf{OfferState}(stock, lifecycle, validFrom, validTo) is one version of an offer, attached by \textsc{has\_state} and chained by \textsc{previous}; a new version opens whenever an observed value differs from the current one, its interval runs to the update time of the next version, and the latest version is open. A \textbf{Query}(text, nSearches, nDays) is one node per normalized query string, connected to the parts returned for it by \textsc{searched\_for}(day, n, exact), where $n$ counts the records of that day.

The graph is built in three stages. Each monthly directory is parsed in
parallel into normalized tables of parts, queries, distributors, offer states
and daily demand, with consecutive identical offer versions collapsed. Monthly tables are unified by normalized MPN, normalized query text and distributor identifier, with one node per entity and offer-state chains linked across months. The unified tables are bulk-loaded into a graph database for querying and inspection, and the method reads them directly.
The graph is updated by appending. A new monthly delivery is parsed into
the same tables, and its parts, queries, distributors and offers are matched
to existing nodes by their keys or added as new nodes. For each offer, an
incoming observation is compared with the open version: if stock and lifecycle
are unchanged no node is added, and otherwise the open version is closed at
the update time and a new version is opened and linked to it by
\textsc{previous}. The searches of each new day are added as
\textsc{searched\_for} edges. Closed versions and past edges are not modified,
so the state of the graph on any earlier day can be read from the validity
intervals.
Table~\ref{tab:graphsize} gives the size of the resulting graph. The offer
states and daily demand edges carry the temporal signal, while the static
support is more than an order of magnitude smaller. An offer has a median of
five state versions over the period and a mean of about nineteen, so the graph
is dominated by a minority of offers with long chains of short-lived states.

\begin{table}[t]
\centering\footnotesize
\caption{{Size of the constructed graph.}}
\label{tab:graphsize}
\begin{tabular}{@{}lr@{\hspace{2em}}lr@{}}
\toprule
Nodes & & Relationships & \\
\midrule
Manufacturer & 37,601 & \textsc{made\_by} & 39.11\,M \\
Part & 39.21\,M & \textsc{offers} & 104.34\,M \\
Distributor & 109 & \textsc{for} & 104.34\,M \\
Offer & 104.34\,M & \textsc{has\_state} & 1.95\,B \\
OfferState & 1.95\,B & \textsc{previous} & 1.84\,B \\
Query & 38.43\,M & \textsc{searched\_for} & 5.82\,B \\
\midrule
Total & 2.13\,B & Total & 9.86\,B \\
\bottomrule
\end{tabular}
\end{table}

For reasoning, the intervals and daily edges are read as temporal ground atoms.
Time is discretized into weekly anchors $a$ (Mondays from 22 December 2025 to 20 April 2026) and signed horizons
$\delta$ of 0, $\pm$1, $\pm$2, $\pm$4, $\pm$7, $\pm$14, $\pm$28 and $\pm$56 days.
An atom $r(e, a, \delta)$ states that predicate $r$ holds for entity $e$ on the
target day $a+\delta$, as seen from anchor $a$; for instance
$\textsc{StockGE100}(p, a, {+}7)$ states that part $p$ has at least 100 units in
stock across distributors one week after $a$.

The temporal predicates form four
families. \textsc{StockGE} (five thresholds on the stock of a part summed over
distributors), \textsc{OfferStockGE} (five thresholds on the stock of one offer)
and \textsc{Active} (whether the lifecycle code of the offer is an active
status) are states on the target day. An offer state is attributed to day $t$
only if it is the latest state that began by the end of $t$ and its end is no
earlier than seven days before the end of $t$, where a state with no recorded
end is given an end at the last instant of the calendar month in which it began,
so that a listing not reported again in the following month drops out a week
after the month end; an offer without such a state is unobserved on $t$ rather
than inactive or out of stock, and the part stock on $t$ is the sum over the
offers with such a fresh state and is observed only when at least one offer has
one. When a distributor reports several entries for a part at the same time,
their stock is summed and the lifecycle code of the entry with the largest stock
is used. \textsc{Active} is true for active, new or in-production codes, false
for obsolete, discontinued, not-recommended or last-time-buy codes, and
unobserved for unmapped codes. \textsc{DemandGE} (six thresholds) counts searches for a part from the day after the earlier of anchor and target through the later day, inclusive. Only horizon $+7$ gives the week ending at the target. Demand uses fourteen nonzero horizons; state predicates also use horizon 0.

An atom is
observed when every day it describes lies on or before the observation cut, and
latent otherwise. The schema relations and four closed-world predicates over
time constants complete the vocabulary: $\textsc{Gap}_g$ links two horizons of
one anchor whose target days differ by $g$ days, for seven gap sizes;
$\textsc{Mirror}_k$ links the horizons $-k$ and $+k$ of one anchor;
\textsc{Wider} orders horizons of the same sign; and \textsc{InMonth} assigns an
anchor to its calendar month. Atoms are assigned to evidence, validation and
test by the last day they describe, with cuts at 28 February and 31 March 2026;
an atom whose window crosses the first cut is held out or dropped, so no search
or state after 28 February enters the evidence.

\section{Methodology}
\label{sec:method}

\begin{figure*}[t]
  \centering
  \includegraphics[width=1\textwidth]{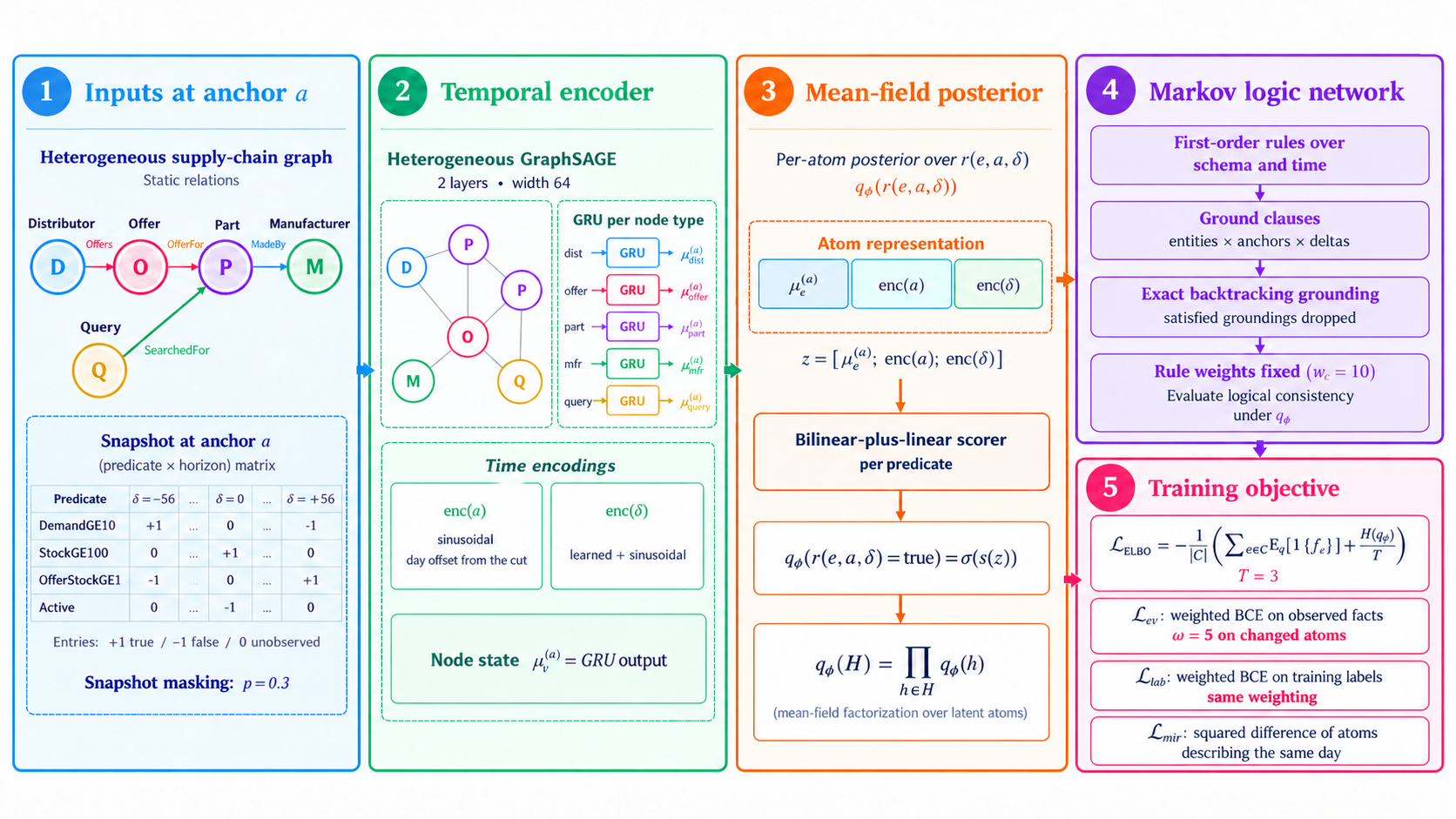}\vspace{-10pt}
  \caption{Overview of the model. (1) The temporal facts of an entity at an anchor form a (predicate $\times$ horizon) snapshot. (2) A snapshot GNN encodes it with the entity's schema neighborhood. (3) A recurrent cell carries the entity state across anchors. (4) Predicate scorers map the state, anchor and horizon encodings to every temporal atom's posterior. (5) Training uses the ground rules, the observed facts and the mirror.
  }
  \label{fig:overview}
\end{figure*}

Let $G = (\mathcal{E}, \mathcal{R}_s, \mathcal{R}_t)$ be the temporal
knowledge graph of Sec.~\ref{sec:kg}, with typed entities $\mathcal{E}$,
fully observed schema relations $\mathcal{R}_s$ and temporal predicates
$\mathcal{R}_t$, whose ground atoms $r(e, a, \delta)$ over anchors
$a \in \mathcal{A}$ and horizons $\delta \in \mathcal{D}$ are defined in
Sec.~\ref{sec:atoms}. Writing $\mathcal{O}$ for the observed and
$\mathcal{H}$ for the latent atoms, the forecasting problem is to estimate
$P\big(r(e, a, \delta) = 1 \mid \mathcal{O}\big)$ for every latent atom,
that is, for every entity, predicate, anchor and horizon jointly. Besides
the sparsity and persistence of Sec.~\ref{sec:data}, one further property
shapes the solution: the same calendar day $t$ is described by every atom
with $a + \delta = t$, so a long-range forecast from an early anchor and a
short-range forecast from a later anchor refer to the same unknown for the
state predicates.

We model the joint distribution of observed and latent atoms as an MLN \cite{richardson2006mln}, since the regularities of
the domain are expressed naturally as weighted first-order rules over the
schema and over time: the state of a part at one anchor conditions its
state at the next, parts of one manufacturer that share queries covary,
and the state described by atoms of the same day from different anchors
must agree. An
MLN with ground clauses $\mathcal{C}$ and weights $w_c$ defines
$P(\mathcal{O}, \mathcal{H}) \propto \exp\big(\sum_{c \in \mathcal{C}} w_c \mathbb{1}_{f_c}\big)$,
and the posterior $P(\mathcal{H} \mid \mathcal{O})$ is intractable at
this scale. We approximate it with a mean-field variational posterior
$q_\phi(\mathcal{H}) = \prod_{h \in \mathcal{H}} q_\phi(h)$ whose
parameters are produced by a graph neural encoder, in the variational
framework introduced by ExpressGNN \cite{zhang2020expressgnn}. That
framework assigns one embedding per constant and treats every ground atom
as an independent variable, so neither the order of anchors nor the
coincidence of target days is represented. The model (Fig.~\ref{fig:overview}) makes time explicit in the grounding, encoder, posterior and objective.

The 317 rules instantiate nine templates over threshold levels and time
gaps. Four examples:
{\footnotesize
\begin{align*}
&\textsc{DemandGE5}(p,a,\delta) \Rightarrow \textsc{DemandGE2}(p,a,\delta),\\
&\textsc{StockGE100}(p,a,\delta_1) \wedge \textsc{Gap}_{7}(\delta_1,\delta_2)\\
&\qquad \Rightarrow \textsc{StockGE100}(p,a,\delta_2),\\
&\textsc{OfferStockGE10}(o,a,\delta) \wedge \textsc{OfferFor}(o,p)\\
&\qquad \Rightarrow \textsc{StockGE10}(p,a,\delta),\\
&\textsc{DemandGE1}(p_1,a,\delta) \wedge \textsc{SearchedFor}(q,p_1)\\
&\qquad \wedge \textsc{SearchedFor}(q,p_2) \Rightarrow \textsc{DemandGE1}(p_2,a,\delta).
\end{align*}}
The templates cover threshold chains (13 rules), the lift from offer stock to
part stock (5), monotonicity of demand over wider windows (6), soft persistence of stock and active status across seven gap sizes in both directions together
with the implication from later stock to earlier active status (161), demand before
the anchor combined with later stock (42), the demand mirror between the
horizons $-k$ and $+k$ of one anchor (84) and shared demand of co-searched
parts (6).

Since the structural predicates of Sec.~\ref{sec:atoms} (\textsc{Gap},
\textsc{Mirror}, \textsc{Wider}, \textsc{InMonth}) and the schema
relations are closed-world, most literals of a rule are decidable at
grounding time. In place of the random instantiation of
\cite{zhang2020expressgnn}, ground clauses are enumerated by an exact
backtracking join that binds the closed-world literals first and the
remaining variables over their domains. Groundings already satisfied by the
evidence are dropped as constants of the objective, groundings whose literals
are all false under the evidence are dropped as violated, and each remaining
ground clause, which contains at least one latent atom, retains only its
latent literals. Reservoir sampling retains a uniform subset of 200,000
candidate clauses from each rule. On the graph of Sec.~\ref{sec:experiments}
all 317 rules reach the cap, yielding 63.4 million clauses, about 2.8 percent
of the 2.26 billion candidates; inclusion probabilities differ across rules despite equal retained counts. Sampling uses the run seed and is cached; ablations share the seed 0 cache.

The encoder (Fig.~\ref{fig:overview}, panels 1 to 3) produces a state
$\mu_e^{(a)}$ for each entity at each anchor. The input of entity $e$ at anchor $a$ is a base
embedding plus a linear projection of the temporal facts observed for $e$
at $a$, arranged as a (predicate $\times$ horizon) matrix with entries
$+1$, $-1$ and $0$ for true, false and unobserved. Two heterogeneous SAGE
layers \cite{hamilton2017inductive} propagate messages along the schema relations in both directions,
giving a per-anchor vector $x_e^{(a)}$. A type-specific GRU cell then
runs over anchors in chronological order,
\[
  \mu_e^{(a)} = \mathrm{GRU}_{\mathrm{type}(e)}\!\big(x_e^{(a)}, \mu_e^{(a^-)}\big),
\]
where $a^-$ denotes the preceding anchor. Anchors without observed facts
therefore receive a state carried forward from history, so the model extrapolates past the last observed week. Snapshot features are
randomly masked during training so that the state does not depend on the
current anchor alone. Time constants are not embedded as entities: an
anchor is represented by a sinusoidal encoding of its day offset from the
observation cut, and a horizon by a learned embedding added to the
sinusoidal encoding of its offset, so that nearby horizons receive nearby
representations and anchors outside the training range are supported
without retraining. When the graph is updated, the new facts of an
entity already in the graph enter the snapshots of the later anchors, and one
forward pass of the trained encoder gives the updated posteriors; entities
that are new to the graph have no base embedding and require retraining.

Each predicate $r$ has a bilinear-plus-linear scorer $s_r$, as in
\cite{zhang2020expressgnn}, applied to the entity state at the atom's anchor and the encodings of its two time constants
(Fig.~\ref{fig:overview}, panel 4),
\[
  q_\phi\big(r(e, a, \delta) = 1\big) = \sigma\!\Big(
    s_r\big([\mu_e^{(a)};\, \mathrm{enc}(a);\, \mathrm{enc}(\delta)]\big)\Big).
\]
The same scorer serves every horizon, so horizon dependence is carried
by $\mathrm{enc}(\delta)$ and by the recurrent state. One forward pass yields the posterior of every atom of
every predicate, threshold and horizon.

We learn $\phi$ by minimizing
(Fig.~\ref{fig:overview}, panel 5)
\begin{equation}
  \mathcal{L} = \mathcal{L}_{\text{ELBO}}
   + \mathcal{L}_{\text{ev}}
   + \mathcal{L}_{\text{lab}}
   + \mathcal{L}_{\text{mir}} ,
  \label{eq:full}
\end{equation}
with the four terms of Eq.~\eqref{eq:full} weighted equally. The first is the negative evidence
lower bound of the MLN,
\begin{equation}
  \mathcal{L}_{\text{ELBO}} = -\frac{1}{|\mathcal{C}|}\Big(
    \sum_{c \in \mathcal{C}} w_c\, \mathbb{E}_{q_\phi}[\mathbb{1}_{f_c}]
    + \frac{1}{T}\, H(q_\phi) \Big),
  \label{eq:elbo}
\end{equation}
with entropy temperature $T$. For a Horn clause the expected satisfaction
has the closed form
$\mathbb{E}_{q}[\mathbb{1}_{f_c}] = 1 - \prod_{\ell \in c} P_q(\ell\ \text{false})$,
evaluated in vectorized form over padded clause tensors. Rule weights
$w_c$ are held fixed at their initial values, 10 for the 24 rules that
encode threshold chains, monotonicity over wider windows and the lift from
offer stock to part stock, and 1 for the remaining 293; learning them by the pseudo-likelihood M-step of \cite{zhang2020expressgnn} is an ablation.

The terms $\mathcal{L}_{\text{ev}}$ and $\mathcal{L}_{\text{lab}}$ are
weighted cross-entropies between $q_\phi$ and, respectively, the observed
temporal facts and the latent atoms of the training period. The weight of
an atom is $\omega > 1$ if its value differs from the last observed label
of the same predicate and entities, the reference defined in
Sec.~\ref{sec:experiments}, and $1$ otherwise. This emphasizes the atoms where the last observation fails, rare in a graph where most states persist, without discarding the persisting atoms.

Eq.~\eqref{eq:elbo} couples two atoms only when some ground clause
mentions both, and the grounded mirror rules relate demand counts in
equal-length windows on opposite sides of an anchor, a soft persistence
assumption. A further term
\[
\mathcal{L}_{\text{mir}} = \operatorname{mean}_{(i,j):\, a_i + \delta_i = a_j + \delta_j}
\big(q_i - q_j\big)^2
\]
pairs the indexed atoms with $a_1 + \delta_1 = a_2 + \delta_2$, which describe
the same calendar day from two anchors. For the state predicates these atoms
describe the same quantity under the common evidence cut, and the term ties
their posteriors. For demand they generally describe different windows: two
positive horizons give nested windows that end on the day, two negative
horizons give nested windows that start after it, and opposite signs give
windows on either side of it, so their probabilities need not agree. The
main runs therefore apply a heuristic regularizer to demand rather than an
identity constraint; a variant that restricts the term to the state
predicates is evaluated in Sec.~\ref{sec:results}. This is the sole temporal regularizer.

Finally, since one entity appears under many anchors, a uniform random
split of latent atoms as in \cite{zhang2020expressgnn} leaks future states
into training, which the split by the last described day of Sec.~\ref{sec:atoms}
prevents. To make the encoder robust to the absence of recent
observations, each training step draws an earlier cut at random, hides
the snapshot entries that describe days after it, carries the last
visible snapshot forward to the later anchors, and applies
$\mathcal{L}_{\text{ev}}$ and $\mathcal{L}_{\text{lab}}$ only to atoms
describing days after that cut; the anchor encoding is taken relative to
the cut. At evaluation the cut is the last observed day.

\section{Experiments}
\label{sec:experiments}

The experiments use a connected slice of the graph of Sec.~\ref{sec:kg}:
the 50,000 parts with the highest co-search degree, their one-hop co-search
neighbors, and every offer, distributor, manufacturer and search query
attached to them (Table~\ref{tab:kg50k}). The cohort and the static
support use the full collection period, while temporal evidence is restricted
to the cut. The cohort is therefore not selected at the cutoff: the
co-search degree that picks the parts also counts searches of the validation
and test months, so the choice of parts, although not their temporal
evidence, uses information from after the cut. The slice favors frequently
searched parts for the same reason, and Sec.~\ref{sec:results} separates
frequently from rarely observed parts. Atoms, anchors, horizons, predicates and cuts are those of Sec.~\ref{sec:atoms}, and clauses are grounded as in Sec.~\ref{sec:method}.

\begin{table}[t]
\centering\footnotesize
\caption{Experimental KG slice.}
\label{tab:kg50k}
\begin{tabular}{@{}lr@{}}
\toprule
Parts & 50,000 \\
Offers & 230,133 \\
Search queries & 141,077 \\
Manufacturers & 1,399 \\
Distributors & 107 \\
Entities (total) & 422,716 \\
Anchors / horizons / months (time constants) & 18 / 15 / 5 \\
\midrule
Predicates (temporal) & 37 (17) \\
Static facts & 753,421 \\
Temporal facts & 178.3\,M \\
First-order rules & 317 \\
Groundings enumerated / with latent literals & 4.0\,B / 2.26\,B \\
Ground clauses kept (200,000 per rule) & 63.4\,M \\
Indexed atoms & 318.2\,M \\
Latent atoms in clauses & 85.1\,M \\
\textsc{Mirror} pairs & 135.1\,M \\
\bottomrule
\end{tabular}
\end{table}

Table~\ref{tab:split50k} gives the split by the last described day. No
fact after the first cut enters the model as evidence, so validation and test
atoms are forecast from the same history, and the validation set only monitors training, since the last-epoch model is evaluated. As most states persist, all metrics are also computed on the reference-error
subset defined below.

\begin{table}[t]
\centering\scriptsize
\caption{Chronological split by the last described day. Positive and reference-error shares refer to the query atoms of each period.}
\label{tab:split50k}
\begin{tabular}{@{}lrrrrr@{}}
\toprule
Split & Target days & Atoms & Share & Positive & Ref.\ err. \\
\midrule
Train & 20 Dec -- 28 Feb & 17.8\,M & 18.6\% & 38.5\% & 0.9\% \\
Valid & 1 -- 31 Mar & 45.6\,M & 47.4\% & 42.6\% & 9.5\% \\
Test & 1 -- 20 Apr & 32.7\,M & 34.0\% & 40.9\% & 9.8\% \\
\midrule
Total & & 96.1\,M & 100\% & 41.2\% & \\
\bottomrule
\end{tabular}
\end{table}

The encoder has two heterogeneous SAGE layers, a type-specific GRU cell and
one base embedding per entity, all of width 64, a one-layer scorer per
predicate and dropout 0.1. Snapshot entries are masked with probability 0.3,
the entropy temperature is $T=3$ and the change weight is $\omega=5$. The training cut is drawn
uniformly at each step from the seven weekly anchors from 12 January to
23 February 2026.
Each step
samples 65,536 ground clauses, evidence or label atoms and \textsc{Mirror}
pairs, and Adam at learning rate $3\times10^{-3}$ runs for 30 epochs without
early stopping. The main model is trained with four seeds; ablations use one, except the state-only mirror variant, which also uses four. Each run uses one GPU, and grounding is cached across runs (Table~\ref{tab:compute}). The time per epoch includes validation, and almost all encoder parameters are the base entity embeddings.

Scores are evaluated by AUC-ROC, AUC-PR (average precision with tied scores entering together at each threshold), accuracy at 0.5 and negative
log-likelihood (NLL), on all test atoms and on the reference-error subset. For
filtered ranking, 5,000 positive test atoms are sampled; the entity, anchor
and horizon are each replaced by up to 200 alternatives, true candidates are
filtered out, and the rank of the true atom is summarized by MRR, mean rank
and Hits@$\{1,3,10\}$, with ties counted against the model. Two deterministic
reference predictors are reported without standard deviations. Persistence
returns the label of the most recent observed fact of the same predicate and
entity whose day lies before the target day, averaged when several fall on one
day, and, when there is no earlier observation, the Laplace-smoothed positive rate
of the predicate over the observed atoms, whose spans end on or before 28
February, so that no validation or test atom enters it; for demand it copies the last observed label regardless
of window length, so a long window may be answered by a short one. Prior
scores every atom by that positive rate.
Because this favors our model on demand, three stronger demand
references are evaluated on all demand atoms of the test split: persistence
restricted to the last observed window of the same length; the raw count of
that window compared with the threshold through a Poisson tail probability;
and the search rate of the last four observed weeks, scaled to the window
length, with the same Poisson tail.

We separately evaluate the atoms whose
label differs from the binary prediction of persistence; for demand the
reference may use a different window length, and atoms without history use the
prior, so this reference-error subset identifies errors of the last observation
and does not exclusively represent temporal transitions. The change weight of
Sec.~\ref{sec:method} uses the same rule. NLL is computed with probabilities
clipped to $[10^{-6}, 1-10^{-6}]$; the NLL of persistence reflects confidently
incorrect hard predictions and soft fallback probabilities and is sensitive to
the clipping threshold. Reference-error detection scores every test atom by the
absolute difference between its probability and the persistence value, with a
reference error as the positive label; for persistence this score is constant
and its AUC-PR is at chance.

We also compare the model with two predictors that query the graph
directly. The first is an agent built on Claude Opus 5.5 at maximum reasoning
effort with two tools, a read-only Cypher interface to a Neo4j copy of the same
slice holding records up to 28 February 2026, and a function that submits one
probability per atom. Its system prompt gives the evaluation date, the schema
and the exact definition of every predicate, including thresholds, offsets and
the freshness rule, and asks for calibrated probabilities. The agent sees no
labels, persistence values or reference-error flags, but is told each queried atom has a known truth value, a conditioning the model does not receive. The static elements come from the
full collection period as they do for the model, and no time-varying fact after
the cut is present. Sessions of 25 atoms were capped at 40 queries of at most
300 rows and 60 model calls; the agent averaged 1.5 queries per atom, and 37 of
53 sessions reached the query cap. The second predictor is a rule with no
learned component: a stock atom is scored by the distance of the last raw stock
value from the threshold, a demand atom by the searches already seen in the
window plus a Poisson forecast from the preceding 28 days, and an \textsc{Active} atom
by its last known value.

All predictors are scored on the same fixed sample of
test atoms (Table~\ref{tab:llmdata}), taken from a shuffled uniform draw of
20,000 test atoms: its first 200 atoms form the uniform sample, and from the
remainder, in draw order, the first 1,000 reference errors and the first 500
other atoms were dealt round robin into 60 groups of 25 before the run. The
agent's budget was exhausted after the first 45 groups in index order, which
hold 760 reference errors and 365 other atoms. The 1,325 atoms involve 1,297
distinct parts. All-atom metrics reweight the two groups to the reference-error share of that draw, 10.03 percent, and
confidence intervals are paired bootstraps over atoms.

Ablations retrain the model with the changes below, retaining other settings, the schedule, cache and selection rule of the main run; all use seed 0 and are scored on the test split at epoch 30. Three variants concern the
training objective: dropping the evidence term; dropping both the evidence
term and the temporal mirror, which leaves the ExpressGNN objective with
label supervision on our encoder; and reducing the change weight from 5 to 1. Three variants switch on joint learning of the rule
weights, alone and combined with the removal of the mirror or the evidence
term. The last two variants alter the architecture: the GRU history encoder is
replaced by the per-anchor embedding without recurrence, and the time encoding
of the horizon is removed. A further variant, trained with four seeds, restricts the mirror term to the eleven state predicates (\textsc{StockGE}, \textsc{OfferStockGE} and \textsc{Active}), which keeps 106.1 of the 135.1 million \textsc{Mirror} pairs and removes the demand pairs for which the term is only a heuristic. No variant removes the rules entirely, so these ablations do not isolate
the incremental contribution of the logical rules beyond the recurrent
encoder.

\begin{table}[t]
\centering\scriptsize
\caption{Compute per training run.}
\label{tab:compute}
\begin{tabular}{@{}lr@{}}
\toprule
GPU & RTX PRO 6000, 96\,GB \\
Peak GPU memory & 84.4\,GB \\
Peak host memory & 155\,GB \\
Setup (cached grounding) & 1.0\,h \\
Time per epoch & 23.7\,min \\
Training (30 epochs) & 11.9\,h \\
Inference over all test atoms & 16.5\,s \\
Final evaluation and ranking & 5\,h \\
Parameters, encoder / scorers & 27.8\,M / 10.1\,M \\
\bottomrule
\end{tabular}
\end{table}

\begin{table}[t]
\centering
\caption{Test atoms for the agent and rule.}
\label{tab:llmdata}
\resizebox{\columnwidth}{!}{%
\begin{tabular}{lrrrl}
\toprule
Sample & Atoms & Ref.\ err. & Ref.\ ok & How chosen \\
\midrule
Uniform          & 200   & 20  & 180 & Uniform random from the test split \\
Error-enriched & 1{,}125 & 760 & 365 & Fixed before the run; 45 of 60 groups completed \\
\midrule
Total                & 1{,}325 & 780 & 545 & \\
\bottomrule
\end{tabular}}
\end{table}

\section{Results}
\label{sec:results}

The epoch with the highest validation AUC-ROC on reference-error queries yields the same test values to within 0.002, so no validation-selected rows are shown. Table~\ref{tab:cls50k} gives query classification per split. Our model exceeds the prior on every split and metric, surpasses
persistence on validation AUC-ROC and AUC-PR, and ties with it on test AUC-PR.
On the test split persistence stays ahead on AUC-ROC and accuracy, because the last observation agrees with more than 90\% of test labels and the test weeks lie further from the observation cut. Our model has less than half the NLL of persistence on both held-out splits. Persistence NLL reflects confidently wrong hard predictions and is sensitive to clipping (Sec.~\ref{sec:experiments}); against the prior, test NLL improves from 0.541 to 0.502.

\begin{table}[t]
\centering
\caption{Query classification per temporal split (mean $\pm$ std over four seeds; reference predictors are deterministic). Arrows give the direction of improvement.}
\label{tab:cls50k}
\resizebox{\columnwidth}{!}{%
\begin{tabular}{llcccc}
\toprule
Split & Predictor & AUC-ROC$\uparrow$ & AUC-PR$\uparrow$ & Acc.$\uparrow$ & NLL$\downarrow$ \\
\midrule
Train & Our model        & 0.994$\pm$.001 & 0.990$\pm$.001 & 0.967$\pm$.001 & 0.092$\pm$.005 \\
      & Persistence & \textbf{1.000}          & \textbf{1.000}          & \textbf{0.991}          & \textbf{0.018} \\
      & Prior       & 0.794          & 0.689          & 0.731          & 0.524 \\
\midrule
Valid & Our model        & \textbf{0.930}$\pm$.003 & \textbf{0.908}$\pm$.004 & 0.811$\pm$.006 & \textbf{0.398}$\pm$.010 \\
      & Persistence & 0.901          & 0.856          & \textbf{0.905}          & 1.246 \\
      & Prior       & 0.785          & 0.710          & 0.712          & 0.550 \\
\midrule
Test  & Our model        & 0.885$\pm$.005 & \textbf{0.848}$\pm$.006 & 0.753$\pm$.002 & \textbf{0.502}$\pm$.007 \\
      & Persistence & \textbf{0.898}          & \textbf{0.848}          & \textbf{0.902}          & 1.245 \\
      & Prior       & 0.785          & 0.702          & 0.717          & 0.541 \\
\bottomrule
\end{tabular}}
\end{table}

Table~\ref{tab:fam50k} decomposes test performance by predicate family. The part-level families \textsc{DemandGE} and \textsc{StockGE}, whose evidence is pooled over many edges of the same part, are predicted most reliably; \textsc{OfferStockGE}, the majority of test queries, is the hardest since each offer carries only its own history, and \textsc{Active} reaches a high AUC-PR because most offers carry an active code, so its positive rate is high. Ranking exceeds persistence in every family.

Table~\ref{tab:demref} compares our model with the demand references of
Sec.~\ref{sec:experiments} on all 6.0 million demand atoms of the test split.
Matching the window length raises the AUC-ROC of persistence from 0.719 to
0.887, which remains below our model, so part of the margin in
Table~\ref{tab:fam50k} came from the mismatched window and part did not. The
two Poisson references, which use the raw counts, exceed our model on AUC-ROC,
AUC-PR and accuracy, while our model keeps the lowest NLL. On demand, the
advantage of our model over simple count rules is therefore in NLL and
not in discrimination.

Grouping the test atoms by how often their part is observed gives the
same picture of where the model helps. From the rarest to the most frequently
observed quartile of parts, the AUC-ROC of our model is 0.839, 0.874, 0.894
and 0.900, against 0.833, 0.883, 0.899 and 0.925 for persistence. Our model is
ahead only for rarely observed parts, where the last observation is old, and
persistence gains more from frequent observation.

For the rule example of Sec.~\ref{sec:method}, we compare the posteriors
of all 231 million pairs of adjacent thresholds of the same entity, anchor and
horizon. Averaged over the four seeds, 84.4\% of the pairs are in the order
required by the rule, 94.4\% are in order or violate it by at most 0.01, and
97.9\% by at most 0.05. The rules thus enter as soft penalties: the order
holds for most pairs but is not guaranteed, and since no variant is trained
without rules, this share is not attributed to them.

\begin{table}[t]
\centering\footnotesize\setlength{\tabcolsep}{4pt}
\caption{Demand references on all 6.0 million \textsc{DemandGE} test atoms. Our model: seed means; references are deterministic.}
\label{tab:demref}
\begin{tabular}{lcccc}
\toprule
Predictor & AUC-ROC$\uparrow$ & AUC-PR$\uparrow$ & Acc.$\uparrow$ & NLL$\downarrow$ \\
\midrule
Our model & 0.936 & 0.949 & 0.842 & \textbf{0.363} \\
Persistence (last window) & 0.719 & 0.724 & 0.700 & 4.14 \\
Persistence (same length) & 0.887 & 0.876 & 0.883 & 1.62 \\
Same length + Poisson & 0.945 & 0.957 & 0.883 & 0.744 \\
Four-week rate + Poisson & \textbf{0.961} & \textbf{0.969} & \textbf{0.897} & 0.408 \\
\bottomrule
\end{tabular}
\end{table}

Table~\ref{tab:chg50k} restricts the evaluation to the reference-error
subset, the 9.8\% of test atoms on which persistence is wrong by construction. Our model separates the atoms that turn out true from
those that turn out false with an AUC-ROC of 0.640, and it detects reference errors among all test atoms with an AUC-PR of 0.325 against a base rate of 0.098. Some atoms enter this subset without an actual
transition, through a different demand window or the prior fallback
(Sec.~\ref{sec:experiments}). The gain over persistence is thus concentrated on
the atoms where the last observation fails, at a small loss of discrimination on
the persisting majority. In filtered ranking (Table~\ref{tab:rank50k}) our model doubles the MRR of persistence and raises Hits@10 by a third, with the largest gain at the top of the ranking. Entity corruptions are harder than value or time corruptions but still exceed persistence by more than a factor of two. The prior has a Hits@1 of zero because it scores all atoms of a predicate identically.

\begin{table}[t]
\centering
\caption{Test performance per family (seed means); persistence ranking in parentheses.}
\label{tab:fam50k}
\resizebox{\columnwidth}{!}{%
\begin{tabular}{lrcccc}
\toprule
Family & Queries & AUC-ROC$\uparrow$ & AUC-PR$\uparrow$ & MRR$\uparrow$ & Hits@10$\uparrow$ \\
\midrule
DemandGE     &  6.0\,M & 0.936 & 0.949 & \textbf{0.396} (0.170) & \textbf{0.796} (0.442) \\
StockGE      &  4.9\,M & 0.903 & 0.915 & \textbf{0.357} (0.164) & \textbf{0.771} (0.417) \\
OfferStockGE & 20.6\,M & 0.869 & 0.746 & \textbf{0.303} (0.150) & \textbf{0.768} (0.401) \\
Active       &  1.2\,M & 0.913 & 0.983 & \textbf{0.332} (0.214) & \textbf{0.715} (0.675) \\
\bottomrule
\end{tabular}}
\end{table}
\begin{table}[t]
\centering\footnotesize
\caption{Reference-error subset: test atoms whose label differs from the binary persistence prediction (9.8\% of the split, positive rate 71.9\%). Det.\ PR: reference-error detection AUC-PR (Sec.~\ref{sec:experiments}), base rate 0.098, at chance for persistence.
Mean $\pm$ std over four seeds; reference predictors are deterministic.}
\label{tab:chg50k}
\begin{tabular}{lccc}
\toprule
Predictor & AUC-ROC$\uparrow$ & AUC-PR$\uparrow$ & Det.\ PR$\uparrow$ \\
\midrule
Our model        & \textbf{0.640}$\pm$.007 & \textbf{0.846}$\pm$.003 & \textbf{0.325}$\pm$.005 \\
Persistence & 0.000 & 0.660 & chance \\
Prior       & 0.377 & 0.666 & 0.143 \\
\bottomrule
\end{tabular}
\end{table}

\begin{table}[t]
\centering
\caption{Filtered ranking on test queries.}
\label{tab:rank50k}
\resizebox{\columnwidth}{!}{%
\begin{tabular}{lccccc}
\toprule
Scorer & MRR$\uparrow$ & Hits@1$\uparrow$ & Hits@3$\uparrow$ & Hits@10$\uparrow$ & Entity MRR$\uparrow$ \\
\midrule
Our model        & \textbf{0.340}$\pm$.003 & \textbf{0.190}$\pm$.007 & \textbf{0.366}$\pm$.006 & \textbf{0.772}$\pm$.004 & \textbf{0.218}$\pm$.010 \\
Persistence & 0.163 & 0.022 & 0.184 & 0.434 & 0.085 \\
Prior       & 0.119 & 0.000 & 0.137 & 0.359 & 0.005 \\
\bottomrule
\end{tabular}}
\end{table}

\begin{figure}[t]
\centering
\includegraphics[width=\columnwidth,trim=0 100.9 0 0,clip]{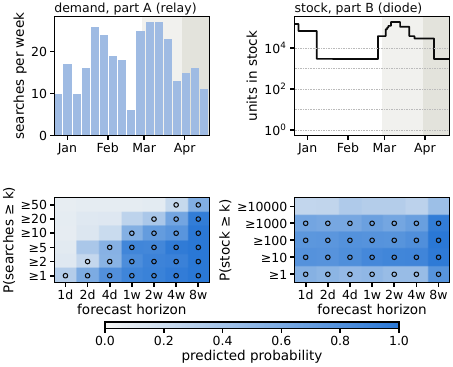}\\[1pt]
\includegraphics[width=\columnwidth,trim=0 0 0 89.6,clip]{figs/FINAL_fig_forecast.pdf}
\caption{Forecasts for two example parts. Top: observed weekly search counts (part A, left) and stock level (part B, right); shaded regions mark the validation and test months. Bottom: posterior probability per threshold $k$ and horizon; hollow circles mark true atoms. Each column uses the latest test anchor for its horizon: 13 April for 1, 2 and 4 days and 1 week, 6 April for 2 weeks, 23 March for 4 weeks and 23 February for 8 weeks. Evidence ends on 28 February, so anchors fix target days and demand windows but not the available information; leads from the last fact are 45 to 51 days for every column. The horizontal axis is thus the horizon of the atom, that is the window length for demand and the offset from anchor to target day for stock, and not the lead time, which is nearly the same in all columns.}

\label{fig:forecast}
\end{figure}

Fig.~\ref{fig:forecast} illustrates the forecasts for two parts, a relay and a diode, with the anchor and target dates of each horizon given in the caption. A dark cell with a circle or a light cell without one is a correct forecast. For the relay, the low thresholds receive high probability at horizons of one week and longer and the high thresholds near zero at short horizons, and the two undecided cells are the ones closest to the realized count. For the diode, whose stock fell from about 30\,000 to about 3\,000 units during the test month, every threshold up to 1\,000 is predicted true and the 10\,000 threshold false at all horizons.

\begin{figure}[t]
\centering
\includegraphics[width=\columnwidth]{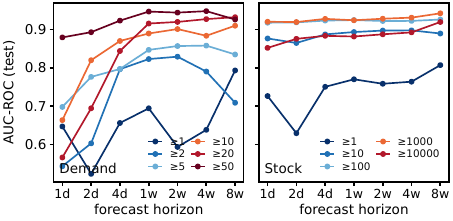}
\caption{Test AUC-ROC per threshold and forecast horizon. The horizon is the window length of a demand atom or the offset from anchor to target day of a stock atom, not the lead from the evidence: evidence ends on 28 February, 32 to 51 days before the test target days, at every horizon.}
\label{fig:horizon}
\end{figure}

Fig.~\ref{fig:horizon} gives the test AUC-ROC per threshold and horizon, where the horizon is the window length of the atom rather than the distance from the evidence. Demand accuracy increases with the horizon, from 0.57 at one day to 0.93 at eight weeks for at least 20 searches, because a longer window is a less noisy count, whereas one- and two-day windows are dominated by whether any search occurs; the lowest threshold of one search remains the hardest at every horizon. Stock atoms are point values and vary little with the horizon: thresholds of 100 and above stay between 0.85 and 0.94, and only the threshold of one unit, which separates stocked from unstocked parts, is lower.

\begin{table*}[t]
\centering

\caption{All predictors on the 1{,}325 sampled test atoms of Table~\ref{tab:llmdata} (780 reference errors, 545 others). Our model is the seed-0 run. Err.: atoms on which persistence is wrong; Ok: atoms on which it is right; starred columns reweight the two groups to a reference-error share of 10.03\%. The detection score of persistence is constant, so its value is chance.}
\label{tab:llm}
\footnotesize
\setlength{\tabcolsep}{4pt}
\begin{tabular}{lcccccccc}
\toprule
Predictor & Err.\ ROC$\uparrow$ & Err.\ PR$\uparrow$ & Err.\ NLL$\downarrow$ & Ok Acc$\uparrow$ & Ok NLL$\downarrow$ & All ROC$^*\uparrow$ & All NLL$^*\downarrow$ & Det.~PR$^*\uparrow$ \\
\midrule
Our model & \textbf{0.614} & \textbf{0.833} & \textbf{0.939} & 0.785 & 0.417 & 0.910 & 0.469 & 0.364 \\
LLM agent & 0.549 & 0.794 & 1.029 & 0.983 & 0.066 & \textbf{0.984} & \textbf{0.163} & \textbf{0.726} \\
Raw-value rule & 0.508 & 0.785 & 2.133 & 0.976 & 0.097 & 0.965 & 0.301 & 0.607 \\
Persistence & 0.000 & 0.658 & 12.633 & \textbf{1.000} & \textbf{0.012} & 0.896 & 1.278 & chance \\
Prior & 0.357 & 0.656 & 1.034 & 0.780 & 0.462 & 0.809 & 0.519 & 0.167 \\
\bottomrule
\end{tabular}
\end{table*}

\begin{table}[t]
\centering
\caption{Reference-error atoms by family (ROC$\uparrow$ / NLL$\downarrow$).}
\label{tab:llmfam}
\resizebox{\columnwidth}{!}{%
\begin{tabular}{lrccc}
\toprule
Family & $n$ & Agent & Our model & Raw rule \\
\midrule
Active       &   5 & -- / 1.334    & -- / \textbf{0.910}    & -- / 2.430 \\
DemandGE     & 433 & 0.785 / 0.688 & \textbf{0.826} / \textbf{0.562} & 0.780 / 1.130 \\
OfferStockGE & 218 & \textbf{0.329} / \textbf{1.446} & 0.190 / 1.564 & 0.099 / 3.363 \\
StockGE      & 124 & 0.201 / 1.476 & 0.144 / \textbf{1.162} & \textbf{0.324} / 3.459 \\
\bottomrule
\end{tabular}}
\end{table}

Table~\ref{tab:llm} scores every predictor on the 1{,}325 sampled test atoms of Table~\ref{tab:llmdata}, with columns defined in its caption; ranking is not compared because the sample contains only fifteen ranking items. On the reference errors our model has the highest AUC-ROC and AUC-PR and the
lowest NLL. A paired bootstrap over atoms puts its AUC-ROC advantage at $+0.023$
to $+0.108$ over the agent and $+0.070$ to $+0.143$ over the rule, and its
AUC-PR advantage at $+0.016$ to $+0.062$ and $+0.027$ to $+0.069$ (95\%
intervals); the upper end of the NLL interval against the agent rounds to zero.
Table~\ref{tab:llmfam} locates the gain in demand atoms, over half the reference errors, whose window counts are predictable from recent searches for the part and its neighbors. Stock atoms in this subset have below-chance AUC-ROC for every predictor.

On the atoms where the reference is right the ordering reverses. The agent
reads the last recorded state from the graph and keeps its probability close
to it, adjusting for how volatile the offer and its distributor have been, and
the rule hedges the last value only slightly, so both approach persistence on accuracy and NLL and outperform our model on the reweighted metrics, including reference-error detection, where the agent's AUC-PR is nearly twice ours. Our model favors the alternative value, losing accuracy on the roughly nine atoms in ten where the reference is right and gaining where it is wrong. The agent is also told that each queried atom has a known truth value, unlike the model (Sec.~\ref{sec:experiments}). The agent needs a model call and averages 1.5 Cypher queries per atom; our model scores all test atoms in one pass.

\begin{table*}[t]
\centering
\caption{Ablation on the test split (seed 0, epoch 30). E: evidence term, M: learned rule weights, T: mirror; Det.~PR: detection AUC-PR.}
\label{tab:abl50k}
\resizebox{\textwidth}{!}{%
\begin{tabular}{lcccccccccccc}
\toprule
 & \multicolumn{4}{c}{All queries} & \multicolumn{4}{c}{Ref.\ err.} & & \multicolumn{3}{c}{Ranking} \\
\cmidrule(lr){2-5}\cmidrule(lr){6-9}\cmidrule(lr){11-13}
Variant & ROC$\uparrow$ & PR$\uparrow$ & Acc$\uparrow$ & NLL$\downarrow$ & ROC$\uparrow$ & PR$\uparrow$ & Acc$\uparrow$ & NLL$\downarrow$ & Det.~PR$\uparrow$ & MRR$\uparrow$ & H@10$\uparrow$ & MR$\downarrow$ \\
\midrule
Our model (E+T, change weight 5)             & \textbf{0.892} & \textbf{0.855} & 0.755 & 0.493 & 0.628 & 0.841 & 0.559 & 0.925 & 0.331 & 0.335 & 0.766 & \textbf{12.5} \\
\midrule
w/o evidence term (T) & 0.886 & 0.849 & 0.751 & 0.500 & 0.659 & \textbf{0.856} & 0.576 & 0.859 & \textbf{0.335} & \textbf{0.351} & \textbf{0.777} & 13.4 \\
w/o evidence and mirror (ELBO + labels), as ExpressGNN & 0.835 & 0.790 & 0.734 & 0.574 & 0.549 & 0.784 & 0.467 & 1.191 & 0.294 & 0.282 & 0.723 & 20.1 \\
\midrule
+ rule-weight learning (E+M+T)          & 0.839 & 0.746 & 0.738 & 0.547 & 0.416 & 0.656 & 0.425 & 1.501 & 0.163 & 0.286 & 0.703 & 21.4 \\
\quad w/o temporal mirror (E+M)         & 0.753 & 0.643 & 0.705 & 1.537 & 0.476 & 0.676 & 0.498 & 2.254 & 0.135 & 0.292 & 0.677 & 24.9 \\
\quad w/o evidence term (M+T)           & 0.750 & 0.632 & 0.691 & 0.710 & 0.443 & 0.689 & 0.453 & 1.065 & 0.132 & 0.307 & 0.722 & 22.3 \\
\midrule
change weight 1                         & 0.883 & 0.851 & 0.756 & 0.495 & 0.649 & 0.847 & 0.568 & 0.882 & 0.325 & 0.343 & 0.767 & 13.5 \\
w/o GRU history encoder                 & 0.844 & 0.757 & \textbf{0.763} & \textbf{0.475} & 0.399 & 0.649 & 0.428 & 0.909 & 0.193 & 0.122 & 0.364 & 64.4 \\
w/o time encoding                       & 0.874 & 0.837 & 0.745 & 0.511 & \textbf{0.666} & \textbf{0.856} & \textbf{0.583} & \textbf{0.848} & 0.322 & 0.347 & 0.759 & 15.4 \\
\bottomrule
\end{tabular}}
\end{table*}

\begin{table*}[t]
\centering
\caption{Five questions, each one temporal atom, asked of the graph with Cypher and of our model.}
\label{tab:cypher}
\resizebox{\textwidth}{!}{%
\begin{tabular}{@{}p{5.2cm}p{5.6cm}p{3.6cm}p{3.4cm}@{}}
\toprule
Question & Cypher result (last observed state, i.e.\ the persistence answer) & Our model: probability; ground truth & Time, all parts: Cypher / Our model \\
\midrule
\multicolumn{4}{@{}l}{\emph{Questions about a date inside the graph (30 Jan 2026)}} \\
Q1 Will part B (a diode) have at least 10{,}000 units in stock across distributors on 30 Jan 2026? & Distributor 1: 66{,}000 (19 Dec); distributor 2: 3{,}000 (30 Jan); four other distributors: 0; the largest current value is 3{,}000, the 66{,}000 is 42 days old & $P=0.37$; ground truth: no & 43.1~s for 6 anchors and 7 horizons (2{,}062{,}185 part-date values) / 16.5~s, all \\
Q2 Will part A (a relay) be searched at least 10 times between 27 and 30 Jan 2026? & Daily searches 27, 28, 29, 30 Jan: 10, 17, 11, 3 & $P=0.76$; ground truth: yes & 55.7~s for 6 anchors and 7 horizons (2{,}080{,}722 part-window values) / 16.5~s, all \\
\midrule
\multicolumn{4}{@{}l}{\emph{Questions about a date outside the graph (anchor 23 Mar 2026, target 20 Apr 2026, evidence up to 28 Feb 2026)}}\\
Q3 Will distributor 2 have part C (a memory device) in stock on 20 Apr 2026? & no fact to retrieve & $P=0.09$; ground truth: no & -- / 16.5~s, all \\
Q4 Will part B have an active lifecycle status at distributor 2 on 20 Apr 2026?& no fact to retrieve & $P=0.63$; ground truth: yes & -- / 16.5~s, all \\
Q5 Will part D (a memory device) be out of stock at every distributor on 20 Apr 2026? & no fact to retrieve & $P=0.71$; ground truth: yes & -- / 16.5~s, all \\
\bottomrule
\end{tabular}}
\end{table*}

Table~\ref{tab:abl50k} gives the ablation results. The recurrent cell is the decisive component: without the GRU, ranking collapses and the reference-error queries are scored close to chance, while accuracy at 0.5 is the highest in the table because the posteriors follow the persisting majority. Removing both the evidence term and the temporal mirror costs about 0.06 AUC-ROC overall and 0.08 on reference-error queries, with a matching rise in NLL. The evidence term alone, the time encoding and the change weight each move the overall metrics by less than one point and trade a small gain on all queries against a small loss on the reference-error subset. Learning the rule weights jointly with the posterior is harmful: every metric falls, AUC-ROC on reference-error queries drops below 0.5, and the variants with learned weights also have the highest NLL.
Restricting the mirror to the state predicates does not improve the
model. Over four seeds, the overall test AUC-ROC is 0.882$\pm$.007 against
0.885$\pm$.005 and the NLL 0.522$\pm$.009 against 0.502$\pm$.007. The state
families change little (\textsc{OfferStockGE} 0.880 against 0.869,
\textsc{StockGE} 0.910 against 0.903, \textsc{Active} 0.912 against 0.913),
while demand AUC-ROC falls from 0.936 to 0.886, AUC-ROC on reference-error
queries from 0.640 to 0.532$\pm$.014 and MRR from 0.340 to 0.327. Although the
mirror is not an identity for demand windows, it acts as a useful regularizer
for them, and the main model keeps it on all predicates.

Table~\ref{tab:cypher} poses five questions, each a single temporal atom, to
the graph and to our model. For the first two, which concern dates inside the
training period, Cypher returns raw records: for the diode, a sum over the
last offer states would exceed the threshold although the ground truth is no,
while our model assigns it a probability below one half; for the relay, our
model assigns a high probability to the threshold that was reached. These
probabilities are in-sample and illustrate the comparison rather than measure
accuracy. The remaining three questions concern a date seven weeks after the last fact in the graph, so Cypher has nothing to retrieve, whereas our model
answers them in the same pass, with the probability on the side of the ground
truth in all three cases. In cost, one inference pass of our model gives the
posterior of every atom in less time than either Cypher query alone, whose
answer amounts to the persistence predictor and exists for past dates only.
This comparison concerns the marginal cost of a question, since the model has
a one-off setup and training cost (Table~\ref{tab:compute}), and the Cypher
times are for a single hand-written query on an untuned server, which tuning or materialized aggregates by part would reduce.

\section{Conclusion}
\label{sec:conclusion}

This work built a temporal knowledge graph from sparse, irregularly and
partially observed electronics supply chain data and forecast its unobserved
atoms with a GNN trained under temporal first-order rules. A single inference
pass scores every test atom of three held-out weeks in seconds. The last
observation remains a strong reference on discrimination over all test atoms,
while on the atoms it gets wrong our model is better than an LLM agent that
queries the graph and than a raw value rule, and the agent is more accurate
where the reference is right. The contribution is thus a scalable
probabilistic forecast of every atom, with a lower negative log-likelihood
than persistence and the prior over all test atoms, and not a higher accuracy
than the last observation; for demand, count-based Poisson references
discriminate better than our model. The evaluation covers one data source over four months, one observation cut and an experimental slice of the graph whose parts were selected with search counts of the full period; the agent comparison uses a sample of 1,325 test atoms. Data volume and sparse, incomplete observations remain challenges. Future work includes a rule-free recurrent
baseline with multi-seed ablations, rolling observation cuts with cohorts selected at each cutoff without future
information, evaluation on the full graph and a broader comparison with query
agents. This
study provides an initial industrial evaluation of probabilistic forecasting
for component availability and search demand on a large temporal knowledge
graph, intended to help engineers assess the sourcing risks of the components
they design in.

\section*{Acknowledgments}
The authors used Claude (Anthropic) and ChatGPT (OpenAI) to revise the wording of the manuscript and to help prepare Figs.~\ref{fig:ontology} and~\ref{fig:overview}. The authors verified all content and take responsibility for it.


\begin{thebibliography}{10}
\bibitem{sandborn2011forecasting}
P.~Sandborn, V.~Prabhakar, and O.~Ahmad, ``Forecasting electronic part
  procurement lifetimes to enable the management of {DMSMS} obsolescence,''
  \emph{Microelectronics Reliability}, vol.~51, no.~2, pp. 392--399, 2011.

\bibitem{jennings2016forecasting}
C.~Jennings, D.~Wu, and J.~Terpenny, ``Forecasting obsolescence risk and
  product life cycle with machine learning,'' \emph{IEEE Transactions on
  Components, Packaging and Manufacturing Technology}, vol.~6, no.~9, pp.
  1428--1439, 2016.

\bibitem{hamilton2017inductive}
W.~L. Hamilton, R.~Ying, and J.~Leskovec, ``Inductive representation learning
  on large graphs,'' in \emph{Advances in Neural Information Processing Systems
  (NeurIPS)}, vol.~30, 2017, pp. 1024--1034.

\bibitem{brintrup2018predicting}
A.~Brintrup, P.~Wichmann, P.~Woodall, D.~McFarlane, E.~Nicks, and W.~Krechel,
  ``Predicting hidden links in supply networks,'' \emph{Complexity}, vol. 2018,
  Art. no. 9104387, 2018.

\bibitem{kosasih2022machine}
E.~E. Kosasih and A.~Brintrup, ``A machine learning approach for predicting
  hidden links in supply chain with graph neural networks,'' \emph{International
  Journal of Production Research}, vol.~60, no.~17, pp. 5380--5393, 2022.

\bibitem{boschee2015}
E.~Boschee, J.~Lautenschlager, S.~O'Brien, S.~Shellman, J.~Starz, and
  M.~Ward, ``{ICEWS} coded event data,'' Harvard Dataverse, 2015, doi:
  10.7910/DVN/28075.

\bibitem{leetaru2013}
K.~Leetaru and P.~A. Schrodt, ``{GDELT}: Global data on events, location and
  tone, 1979--2012,'' in \emph{Int. Studies Association Annual Convention},
  San Francisco, CA, USA, 2013.

\bibitem{huang2023}
S.~Huang, F.~Poursafaei, J.~Danovitch, M.~Fey, W.~Hu, E.~Rossi, J.~Leskovec,
  M.~Bronstein, G.~Rabusseau, and R.~Rabbany, ``Temporal graph benchmark for
  machine learning on temporal graphs,'' in \emph{Advances in Neural
  Information Processing Systems (NeurIPS), Datasets and Benchmarks Track},
  2023.

\bibitem{wasi2024supplygraph}
A.~T. Wasi, M.~S. Islam, and A.~R. Akib, ``{SupplyGraph}: A benchmark dataset
  for supply chain planning using graph neural networks,'' in \emph{Proc. AAAI
  Workshop on Graphs and More Complex Structures for Learning and Reasoning},
  2024, arXiv:2401.15299.

\bibitem{wichmann2020extracting}
P.~Wichmann, A.~Brintrup, S.~Baker, P.~Woodall, and D.~McFarlane, ``Extracting
  supply chain maps from news articles using deep neural networks,''
  \emph{International Journal of Production Research}, vol.~58, no.~17, pp.
  5320--5336, 2020.

\bibitem{liu2023knowledge}
Y.~Liu, B.~He, M.~Hildebrandt, M.~Buchner, D.~Inzko,
R.~Wernert, E.~Weigel, D.~Beyer, M.~Berbalk, and V.~Tresp,
``A knowledge graph perspective on supply chain resilience,''
in \emph{Proc. 2nd Int. Workshop on Linked Data-driven
Resilience Research (D2R2), co-located with ESWC 2023},
ser. CEUR Workshop Proceedings, 2023, arXiv:2305.08506.

\bibitem{elhambakhsh2025domain}
F. Elhambakhsh, D. Grandi, and H. Ko,
``A domain adaptation of large language models for classifying mechanical assembly components,''
in \textit{Proc. ASME International Design Engineering Technical Conferences and Computers and Information in Engineering Conference (IDETC/CIE)},
vol. 89213, Art. no. V02BT02A033, 2025.
\bibitem{aziz2021data}
A.~Aziz, E.~E. Kosasih, R.-R. Griffiths, and A.~Brintrup, ``Data
  considerations in graph representation learning for supply chain networks,''
  in \emph{ICML Workshop on Machine Learning for Data: Automated Creation,
  Privacy, Bias}, 2021.

\bibitem{kosasih2024towards}
E.~E. Kosasih, F.~Margaroli, S.~Gelli, A.~Aziz, N.~Wildgoose, and A.~Brintrup,
  ``Towards knowledge graph reasoning for supply chain risk management using
  graph neural networks,'' \emph{International Journal of Production Research},
  vol.~62, no.~15, pp. 5596--5612, 2024.

\bibitem{kosasih2025trustworthy}
E.~E. Kosasih and A.~Brintrup, ``Towards trustworthy {AI} for link prediction
  in supply chain knowledge graph: A neurosymbolic reasoning approach,''
  \emph{International Journal of Production Research}, vol.~63, no.~6, pp.
  2268--2290, 2025.

\bibitem{garciaduran2018learning}
A.~Garc{\'i}a-Dur{\'a}n, S.~Duman{\v{c}}i{\'c}, and M.~Niepert, ``Learning
  sequence encoders for temporal knowledge graph completion,'' in \emph{Proc.
  2018 Conf. Empirical Methods in Natural Language Processing (EMNLP)}, 2018,
  pp. 4816--4821.

\bibitem{jin2020renet}
W.~Jin, M.~Qu, X.~Jin, and X.~Ren, ``Recurrent event network: Autoregressive
  structure inference over temporal knowledge graphs,'' in \emph{Proc. 2020
  Conf. Empirical Methods in Natural Language Processing (EMNLP)}, 2020, pp.
  6669--6683.

\bibitem{herding2025ontology}
R.~Herding, K.~J. Fordyce, R.~J. Milne, and L.~M{\"o}nch, ``An ontology for
  semiconductor supply chain planning,'' \emph{IEEE Transactions on
  Semiconductor Manufacturing}, vol.~38, pp. 554--570, 2025.

\bibitem{ramzy2021knowgraphpm}
N.~Ramzy, S.~Auer, J.~Chamanara, and H.~Ehm, ``{KnowGraph-PM}: A knowledge
  graph based pricing model for semiconductor supply chains,'' in
  \emph{Computer and Information Science 2021---Summer}, ser. Studies in
  Computational Intelligence, vol. 985.\hskip 1em plus 0.5em minus
  0.4em\relax Cham, Switzerland: Springer, 2021, pp. 61--75.

\bibitem{ham2026design}
S.~Ham, K.~Lee, S.~Park, J.~Kang, and H.~Hwang,
``Design and implementation of a semiconductor supply chain knowledge graph
for developing a {GraphRAG}-based question answering model,''
in \emph{Proc. 59th Hawaii Int. Conf. System Sciences (HICSS)},
2026, pp. 4828--4836.

\bibitem{richardson2006mln}
M.~Richardson and P.~Domingos, ``Markov logic networks,'' \emph{Machine
  Learning}, vol.~62, no.~1--2, pp. 107--136, 2006.

\bibitem{zhang2020expressgnn}
Y.~Zhang, X.~Chen, Y.~Yang, A.~Ramamurthy, B.~Li, Y.~Qi, and L.~Song,
  ``Efficient probabilistic logic reasoning with graph neural networks,'' in
  \emph{Proc. Int. Conf. Learning Representations (ICLR)}, 2020.

\bibitem{qu2019plogicnet}
M.~Qu and J.~Tang, ``Probabilistic logic neural networks for reasoning,'' in
  \emph{Advances in Neural Information Processing Systems (NeurIPS)}, vol.~32,
  2019.

\bibitem{yang2017neurallp}
F.~Yang, Z.~Yang, and W.~W. Cohen, ``Differentiable learning of logical rules
  for knowledge base reasoning,'' in \emph{Advances in Neural Information
  Processing Systems (NeurIPS)}, vol.~30, 2017.

\bibitem{meilicke2019anyburl}
C.~Meilicke, M.~W. Chekol, D.~Ruffinelli, and H.~Stuckenschmidt, ``Anytime
  bottom-up rule learning for knowledge graph completion,'' in \emph{Proc. 28th
  Int. Joint Conf. Artificial Intelligence (IJCAI)}, 2019, pp. 3137--3143.

\bibitem{lacroix2020tntcomplex}
T.~Lacroix, G.~Obozinski, and N.~Usunier, ``Tensor decompositions for temporal
  knowledge base completion,'' in \emph{Proc. Int. Conf. Learning
  Representations (ICLR)}, 2020.

\bibitem{li2021regcn}
Z.~Li, X.~Jin, W.~Li, S.~Guan, J.~Guo, H.~Shen, Y.~Wang, and X.~Cheng,
  ``Temporal knowledge graph reasoning based on evolutional representation
  learning,'' in \emph{Proc. 44th Int. ACM SIGIR Conf. Research and Development
  in Information Retrieval}, 2021, pp. 408--417.

\bibitem{liu2022tlogic}
Y.~Liu, Y.~Ma, M.~Hildebrandt, M.~Joblin, and V.~Tresp, ``{TLogic}: Temporal
  logical rules for explainable link forecasting on temporal knowledge
  graphs,'' in \emph{Proc. 36th AAAI Conf. Artificial Intelligence}, 2022, pp.
  4120--4127.

\bibitem{liu2026}
Y.~Liu and M.~Zolghadri, ``What is happening in product obsolescence
  management? {Three} decades of review from 1993 to 2023,'' \emph{Research
  in Engineering Design}, vol.~37, Art. no.~9, 2026.

\bibitem{saad2025}
E.~Saad, A.~Mrabah, M.~Besbes, M.~Zolghadri, V.~Czmil, C.~Baron, and
  V.~Bourgeois, ``Zero-shot learning for obsolescence risk forecasting,''
  arXiv:2506.21240, 2025.

\bibitem{moon2022b}
K.~S. Moon, H.~W. Lee, and H.~Kim, ``Adaptive data selection-based machine
  learning algorithm for prediction of component obsolescence,''
  \emph{Sensors}, vol.~22, no.~20, Art. no. 7982, 2022.

\bibitem{mungo2024}
L.~Mungo, A.~Brintrup, D.~Garlaschelli, and F.~Lafond, ``Reconstructing
  supply networks,'' \emph{Journal of Physics: Complexity}, vol.~5, no.~1,
  Art. no. 012001, 2024.

\bibitem{kose2026}
S.~Ko{\c{s}}e, C.~Diem, E.~Dervic, K.~Friesenbichler, G.~Heiler, J.~Hurt,
  H.~Picatto, and P.~Klimek, ``Reconstructing temporal multi-relational firm
  networks at scale using large language models: The case of the
  semiconductor industry,'' arXiv:2605.15842, 2026.

\end{thebibliography}
\end{document}